Tech Science Press

# 3D Instance Segmentation Using Deep Learning on RGB-D Indoor Data

**Siddiqui Muhammad Yasir[1], Amin Muhammad Sadiq[2] and Hyunsik Ahn[3,*]**

[1]Department of Robot System Engineering, Tongmyong University, Busan, 48520, Korea
[2]Department of Information & Technology, University of Central Punjab, Lahore, Pakistan
[3]Department of Electronics Engineering, Tongmyong University, Busan, 48520, Korea
*Corresponding Author: Hyunsik Ahn. Email: hsahn@tu.ac.kr

**Abstract:** 3D object recognition is a challenging task for intelligent and robot systems in industrial and home indoor environments. It is critical for such systems to recognize and segment the 3D object instances that they encounter on a frequent basis. The computer vision, graphics, and machine learning fields have all given it a lot of attention. Traditionally, 3D segmentation was done with hand-crafted features and designed approaches that didn't achieve acceptable performance and couldn't be generalized to large-scale data. Deep learning approaches have lately become the preferred method for 3D segmentation challenges by their great success in 2D computer vision. However, the task of instance segmentation is currently less explored. In this paper, we propose a novel approach for efficient 3D instance segmentation using red green blue and depth (RGB-D) data based on deep learning. The 2D region based convolutional neural networks (Mask R-CNN) deep learning model with point based rending module is adapted to integrate with depth information to recognize and segment 3D instances of objects. In order to generate 3D point cloud coordinates *(x, y, z)*, segmented 2D pixels *(u, v)* of recognized object regions in the RGB image are merged into *(u, v)* points of the depth image. Moreover, we conducted an experiment and analysis to compare our proposed method from various points of view and distances. The experimentation shows the proposed 3D object recognition and instance segmentation are sufficiently beneficial to support object handling in robotic and intelligent systems.

## 1 Introduction

3D instance segmentation is a challenging and fundamental computer vision problem with applications in robotics, autonomous driving, and augmented reality, industrial and home environments. Significant progress in 2D image segmentation and object detection has been made with the advancement of deep learning. 3D segmentation has traditionally been performed using hand-crafted features and engineered methods that have failed to achieve acceptable accuracy and cannot be universally applied to large datasets [1,2]. Due to their breakthrough in 2D computer vision, deep learning algorithms have lately been the preferred tool for 3D segmentation challenges.

The goal of 3D instance segmentation is to develop computational approaches that can anticipate the fine-grained labels of objects in a 3D scene for a variety of applications. Additionally, compared to 2D segmentation, 3D segmentation provides more detailed and specified information such as point clouds, voxels, and meshes, with 3D coordinates, for the practical applications. Deep learning approaches have recently dominated a number of research domains, including computer vision, speech recognition, and natural language processing. 3D deep learning approaches, on the other hand, nevertheless face a number of unsolved issues. For example, because point clouds are irregular, it is difficult to merge features from RGB and depth channels and translating them to high-resolution voxels imposes a considerable processing overhead [1,2].

The capacity of 2D instance segmentation to mask pixels from a single or multiple objects in an image has the potential to improve perception in a variety of applications. The deep learning-based approach to 2D instance segmentation has attracted much attention. In particular, Convolutional Neural Network (CNN) based deep learning models are providing many advanced results. The common detection and region-based deep learning models known as Fast, Faster, and Mask R-CNN have shown impressive results in 2D instance segmentation. In these approaches, a backbone predicts regions of interest and then segments them into foreground and background instances as masks [2,3]. Numerous researchers proposed 3D scene understanding with 2D segmentation using RGB-D data. However, they are insufficient to support advanced intelligent and robotic systems applications as the robot requires more precise 3D geometric information about objects [4,5].

Moreover, 3D instance segmentation that extracts the 3D areas of objects in the $(x, y, z)$ coordinate system is necessary for multiple industrial areas. It is needed to identify target objects for applications such as robot object handling, situation perception for autonomous vehicles, and vision systems for factory automation [4–6]. With the advent of RGB-D sensors that capture color and depth data, 3D segmentation and 3D object identification may be able to achieve advanced fulfillment. Using RGB-D data, multiple approaches are used to infer bounding boxes in 3D point clouds, transform 2D depth maps to 3D point clouds, and apply machine learning techniques to detect objects in RGB images [5,7–9]. Furthermore, these networks require point clouds as an input, and color information cannot be considered properly [10]. Processing with a point cloud as an input is time expensive and requires a significant amount of processing. As a result, our goal is to solve the problem of preprocessing and lower the high demand for resources for a speedy solution. Therefore, our work requires RGB and depth map data as an input of deep learning processing and later used the out-put of deep learning for 3D instance segmentation [11].

The depth map in RGB-D data provides geometric information about the scene or real environment that can be used to identify foreground objects from their background, allowing for better segmentation accuracy [12,13]. However, this simple framework is powerful enough to reduce post-processing and improve 3D instance segmentation accuracy. Methods for understanding 3D scenes can also distinguish between distinct instances of the same class. In this research, the 3D mask backbone predicts the region and final instance masks. Proposal-based techniques construct final instance masks by first predicting and then improving object suggestions. A novel method for recognizing and segmenting 3D objects from a 3D indoor environment using RGB-D data is proposed in this paper, The Mask R-CNN deep learning model has been adapted for 2D segmentation using RGB data. Furthermore, the Point Based Rendering module is applied to achieve the precise geometric information. In order to achieve 3D instance segmentation, the ($x, y, z$) coordinates are computed from 2D segmented pixels by utilizing RGB-D data. The main contributions to this paper are as follows:

1. A novel 3D instance segmentation method is proposed that integrates 2D object recognition using deep learning with depth data.
2. The integration of 2D segmented RGB data with depth data to achieve 3D instance segmentation for foreground objects and their background information is presented.

The remainder of this work is structured as follows. Section 2 outlines the work that is connected to the suggested strategy. The third section goes through the specifics of the proposed 3D instance segmentation. Section 4 explores into the experimental results of the suggested 3D instance segmentation method. Section 5 concludes with a last statement on future effort.

## 2 Related Work

This section discusses prior findings on object recognition and instance segmentation in the areas of computer vision and robotics.

### 2.1 Deep Learning-Based 2D Instance Recognition

In the computer vision area, object recognition, including classification and segmentation, has been a main challenge for a long time before the appearance of deep learning approaches. Image segmentation is typically used to locate the regions and boundaries of objects in images. Instance segmentation recognizes objects of interest in an image efficiently while also generating a high-quality individual segmentation mask for each object [7,13].

In order to recognize objects in images, researchers utilized deep learning techniques. An extraordinary development has been made in 2D object recognition with the improvement of Convolutional Neural Networks (CNNs) [14]. The convolutional neural network is a multi-layered feed-forward neural network that allows learning of hierarchical features. The design of CNN is made by stacking a few hidden layers sequentially on top of each other. A well-known model for object recognition with bounding boxes called "You Only Look Once" (YOLO) involves a neural network that includes an image as an input and predicts the bounding box and class label [14].

The R-CNN is a neural network defined as "Region-based CNN" The techniques were designed and demonstrated as region-based convolutional neural network (R-CNN), Fast R-CNN, and Faster R-CNN for object classification and object recognition. In general, the R-CNN models might be more accurate, but the YOLO family of models is fast, far faster than the R-CNN, achieving real-time object detection. The selective search methods are implemented in R-CNN to classify region classification proposals with less complexity. The features are extracted in CNN layers and are passed to several binary classifiers to classify the classes in particular regions [11,14]. Fig. 1 R-CNN Model Architecture features hierarchies for accurate object detection and semantic segmentation. The approach involved in R-CNN to classify and recognize objects is relatively simple and straightforward. Fig. 1 explains the "selective search" approach used to suggest candidate regions or border boxes of possible objects.

The Mask R-CNN is a computer vision approach based on CNN that can recognize, classify, and mask object instances in 2D images. Two phases are part of Mask R-CNN. First, the model allows proposals for input images to be made for regions where an object can be created. The second is the estimation of the class objects, which will refine the bounding box and construct the mask at the object level based on the first proposal at the object level [15]. In recent challenges, the Mask RCNN is state-of-the-art in terms of region-based image and instance segmentation regardless of object size. These region-based systems typically anticipate masks on a $28 \times 28$ grid. This is sufficient for small objects, but it generates an unpleasant "blobby" output for large objects, which excessively smooths the fine-level details. An alternate sliding window method, which employs a sophisticated network design to anticipate sharp high-resolution masks for huge objects, falls slightly behind in accuracy. While boosting the accuracy of region-based techniques, a region-based segmentation model integrated with point-based rendering (PointRend) can produce masks with fine-level information. Furthermore, PointRend is used to segment objects and scenes in high-quality RGB data quickly and effectively, comparing various methods for effective rendering by sampling pixel labeling tasks. The approaches are based on iterative subdivision, and the refined results are achieved with the point-based rendering (PointRend) neural network module for regional detection. It performs point-based segmentation predictions in adaptively selected regions. PointRend can be applied to both instance and semantic segmentation tasks. While the general concept can be implemented in a variety of approaches [16], in places where prior approaches have over smoothed the object borders, PointRend produces crisp object boundaries. Instance segmentation with PointRend uses a new point-based feature representation to make predictions at adaptively sampled points on the image. PointRend produces substantially more detailed results and iteratively computes its prediction when used to replace Mask with CNN's default mask head [16]. Each step uses bilinear interpolation in smooth regions and produces better resolution predictions at a small number of adaptively chosen points that are likely to be on object boundaries.

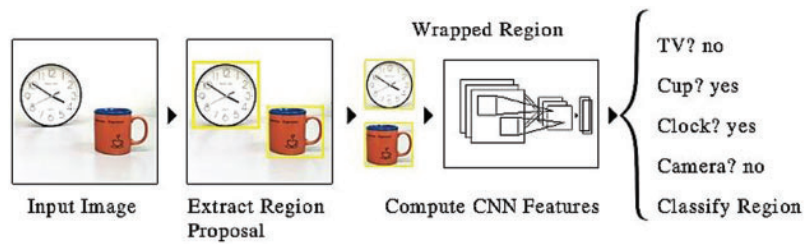

**Figure 1:** General description of convolution neural network (CNN) deep learning model

The PointRend module played an initial role in this research for 2D object recognition and segmentation for targeted object regions in 2D RGB data for further processing.

## 2.2 3D Object Recognition

The 3D object recognition is the ultimate goal of computer vision and can be applied in private areas like homes as well as industrial areas. It extracts meaningful 3D information from various types of data, such as point clouds, RGB-D, scanned data extracted from a range of sensors, and even monocular images [12,16]. The key distinctions between semantic and instance segmentation are as follows. While the semantic segmentation job can be regarded as a classification task with a fixed number of known labels, the object indices for example segmentation are permutation invariant, and the total number of objects is unknown a priori [17]. Currently, there are two primary approaches to instance segmentation: Proposal-based methods seek out intriguing places and then divide them into foreground and background. Proposal-free methods discover a feature embedding space for the pixels in the image [18]. Following that, the pixels are grouped based on their feature vector. In our work, we adopt the first approach since performing instance segmentation in the scene is both robust and fast.

The Mask R-CNN 3D object detector produces 3D box proposals directly by the use of only the raw point cloud as bottom-up data, which is then optimized by the proposed bin-based 3D box regression loss in the canonical coordinate [19]. The 3D bounding boxes are not suitable for robotic systems, particularly in object grasping scenarios. A vehicle shape segmentation approach was proposed by projecting the predicted 2D to 3D confidence maps to be combined [9,20]. The PointIT, a fast tracking framework based on 3D instance segmentation, is proposed recently for object tracking based on 3D instance segmentation. The method first predicts instance mask for each class then uses a convolutional neural network called MobileNet as a primary encoder [20].

Qi et al. [21] used a sequence of multi-layer perceptron (MLPs) and max-pooling to perform feature learning directly on raw point clouds. In the subsequent work [22], hierarchical characteristics are included. Both pieces, however, necessitated extensive processing and resources. The prepressing and input of a point-cloud is a complex process and a burden for speedy instance segmentation.

Similarly, Generative Shape Proposal Network (GSPN) introduces a 3D object proposal network that reconstructs object shapes from shape noisy observations to enforce geometric understanding. GSPN is embedded into a 3d instance segmentation network named as region-based PointNet (R-PointNet) to reject, receive and refine proposals. Step-by-step training of these networks is required, and object proposal refinement requires an complex and costly suppression operation [1,23]. The 3D environment understanding with point cloud became a vital point, since it can advantage numerous applications, such as autonomous driving, augmented, and virtual reality [24]. Current 3D segmentation methods for recognizing objects are based on post-processing on the point clouds and require a lot of processing and these methods are not suitable for real-time and robotic applications [4,25].

However, numerous approaches to object detection utilizing RGB-D data are proposed. Shao et al. merged RGB-D data and clustered the features to produce instance segmentation masks on simulated RGB-D data [26]. The 2D-driven 3D object detection approach has been proposed using synthetic data [27] suitable for real-time and robotic applications for fast processing.

## 3 Deep Learning-Based 3D Instance Segmentation

The new approach for 3D instance segmentation and recognition using RGB-D data is presented in this section. For 3D instance segmentation, the suggested approach revolves around 2D to 3D segmentation methods by using camera calibration together with 2D instance segmentation in 3D space. The proposed approach is divided into three steps; the output of each step is the input of the following step.

Fig. 2 illustrates the overview of the proposed method for 3D instance segmentation from RGB-D data. Fig. 3 demonstrates the masking of each targeted object in an indoor environment. In our approach, we assume that RGB-D data has the same aligned resolution in pairs of a color image and depth data as shown in Figs. 3a and 3b.

First, the 2D Mask R-CNN model has been adapted to mask and recognize the targeted objects. A masking process is required to identify the region of interest (ROI) in RGB for further processing. Later, point based rendering (PointRend) produced crisp object boundaries for more accurate object ROI. The output of the first step is masking, isolating segmented objects in images with distinct color grading between the targeted objects and their background.

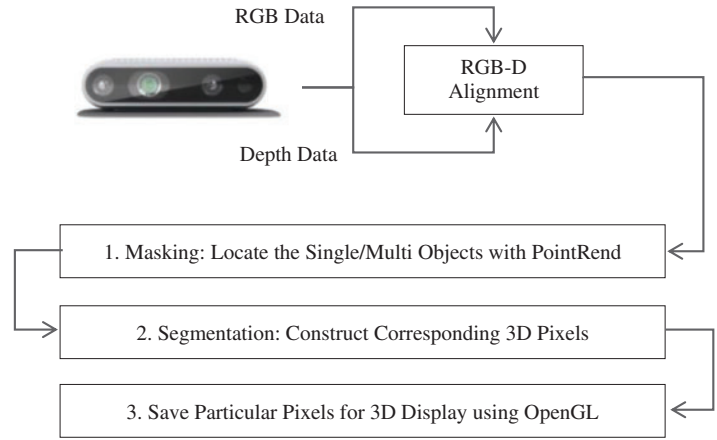

**Figure 2:** Processing pipeline explained for identifying region of interest (ROI) and detaches objects in an indoor environment from their background

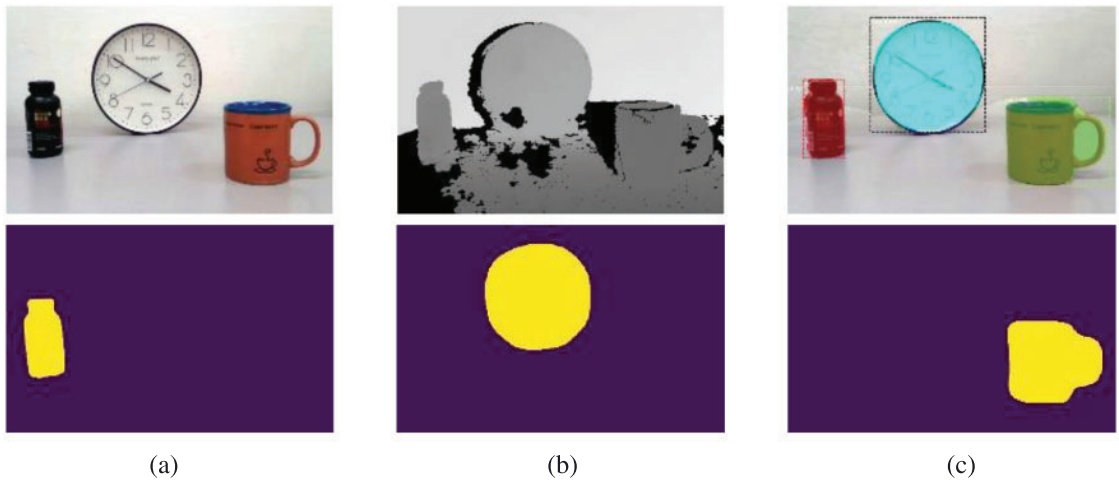

(a)                                          (b)                                          (c)

**Figure 3:** First row shows RGB image (a). The corresponding depth image (b). The refined masked image for multi-objects, identifying the area of each object with their class label in an environment (c). The second row shows distinct masked objects detected from RGB images

Second, after applied to RGB and depth data separately, the technology of contour detection from masks is used to segment the region of interest (ROI) in RGB-D data. Contours are used to connect all continuous points that have the same hue or intensity. There was an issue with unknown or trash information obtained from the sensor in RGB-D data. Therefore we applied a threshold within the range of distance from the sensor to the object to deal with it. As multi-level thresholding is an important consideration in many real-time pattern recognition applications [28]. Third, the ROI detached from their background and converted to 3D pixels with the help of perspective transformation. The following section describes 2D instance segmentation with Mask R-CNN in detail.

The 2D instance segmentation is important to recognize the class and ROI from RGB images. The Mask R-CNN plays the initial role in this research, designed to solve problems like instance segmentation in machine learning or computer vision problems. Faster R-CNN first utilizes a Convolutional Network to extract feature maps from the images [13,28]. These feature maps are then passed through a Region Proposal Network (RPN) which returns the candidate bounding boxes. ROI pooling layer on these candidate bounding boxes can bring all the candidates to the same size. The flattened proposals are passed to a fully connected layer to classify and output the bounding boxes for objects. The Mask R-CNN can determine object class, bounding box, and segment based on pixel points as shown in Fig. 3c. It produces suggestions for regions where objects are placed in an input image. Later, it defines the target class, refines the bounding box, and generates a refined mask with PointRend based on a first-stage proposal at the pixel level of the object.

All the masks are stored in the mask array predicted by PointRend in the form of Boolean labels. Boolean labels always represent 0 s and 1 s, 0 s represent no object at that particular pixel and 1 s represent the existence of the detected object pixel. The shape will represent the number of segmented objects in an image. Each mask is multiplied by the original image to be able to use each segment of the image in a loop. Obtaining the segments from the whole image will reduce the cost of computing as it will not preprocess the whole image, but only the segments.

The pixel-level mask is essential for this research because it targets the exact ROI. The mask of the objects is generated by isolating the individual segmented instances as shown in Fig. 3d. Each mask is established with the isolated area and binary coloring of the ROI with their background information for further processing. In order to achieve 3D pixel values, distinct binary masks are applied to regions of interest on the corresponding depth information.

The next section will explain the process of 3D instance segmentation involving masked images with distinct colors. The 3D instance segmentation is accomplished with the assistance of PointRend refined masks. Fig. 2 demonstrates the process of both phases. The first phase of research focuses on the recognition of objects by masking the region of interest. The second phase of the research is segmentation, which aims at achieving sufficient 3D matching pixels from both RGB-D data.

This section explains how the target area is separated and transformed into three-dimensional information for each object using RGB-D data that is already aligned by the sensor. The 2D key points, defined by adding contours to RGB images, extracted and combined the corresponding depth values from the associated depth and RGB images to obtain the corresponding 3D pixel points. The required target was achieved by marking the masked area in the RGB image with the help of the contour boundary. The area within the boundary was calculated and the targeted RGB-D data pixel values were acquired from the corresponding aligned color and depth images.

The depth values are extracted from the depth image using the pixel locations obtained from the area of interest with the aligned RGB image. Besides, the average depth distance was calculated to ignore the environmental background and to fill the non-targeted pixels with nulls to achieve the best results, as demonstrated in Fig. 4b. Furthermore, the information obtained from RGB-D data was merged for further processing. A 3D environment is drawn once a desired color and depth pixel is generated using the Open3D a computer vision library.

As compared to the other two networks, Mask R-CNN has high accuracy. Even without occlusion, images of several objects are the most complex and impossible to detect. The Mask R-CNN with pertained weights obtained the best results, while the masked region still lacks certain pixels that belong to the objects. Tab. 1 illustrates the object detection accuracy of single and multi-objects.

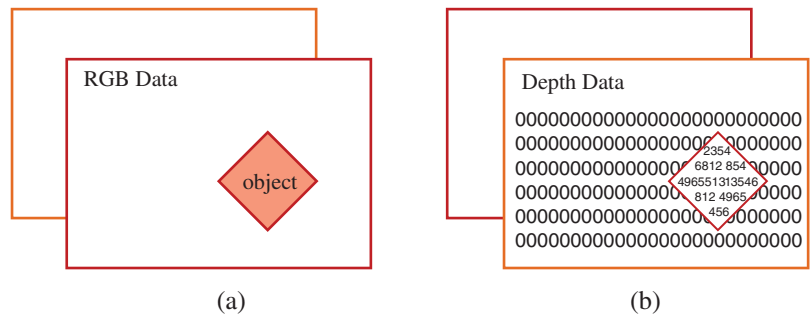

**Figure 4:** RGB data with a masked area of an object (a). Corresponding targeted pixel values of the targeted region from both RGB and Depth images (b). The region of interest (ROI) was differentiated on the base of individual masked object pixels by utilizing the contour methods and applied on both RGB and depth data

**Table 1:** Comparisons with Fast, Faster R-CNN and Mask R-CNN in terms of object detection task using same dataset for all three models. The number shows the sum of accuracy for multiple objects detection for three different models

| Methods | Recognition rate | |
|---|---|---|
| | Single object | Multi objects |
| Fast R-CNN | 90% | 84% |
| Faster R-CNN | 94% | 89% |
| Mask R-CNN | 98% | 92% |

The Mask R-CNN was chosen for single and multi-object detection for experiment purposes due to its accuracy with single and multi-object detection. Moreover, PointRend refined the masked area of interest for more accurate detection.

This subsection describes the proposed process of 3D instance segmentation and recognition of an object as well as multi-objects from the indoor environment. Fig. 4 explains the second phase of segmentation as described in Fig. 2 in our proposed research. The recognition of the object is achieved with PointRend and the instance segmentation for a region of interest (ROI) in RGB-D pixels is achieved with contour technology [29].

Besides, contour technology is used to measure the area of masked pixels. The standard explanation of contours is used to find the markers and measure the region of interest (ROI) for the selected object. Computer vision techniques focus on the use of bottom-up signals to produce images for boxes and regions. Such techniques usually detect contours in images to produce hierarchical segmentation [29].

The regions of this hierarchical segmentation are combined to generate a list of areas with low-level object indices that represent objects in the image.

The architecture of the proposed method is illustrated in Fig. 5 revolves around the two steps of the research. The first step of our approach is to recognize and highlight region of interest (ROI) the target objects, which was accomplished by utilizing the PointRend method. The output of the first phase will be the input of the second phase as a masked RGB image of each object in an indoor environment, illustrated in Fig. 5. The second phase determined the area of each object by applying the contour as explained in Fig. 4. The use of pixel values could be a way of segmenting various objects. As the contrast increases, the pixel values for the objects and the image's background differ. As explained in Fig. 4, 2D three-channel images with high contrast give a clear idea of the region of interest (ROI) and background, explained in Fig. 4. Furthermore, the RGB and the corresponding depth values are acquired by providing the depth image as an input for calculating 3D coordinates with perspective transformation.

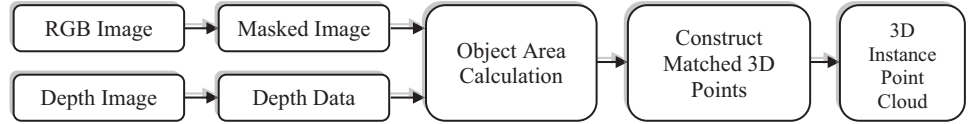

**Figure 5:** The architectre of proposed 3D instance segmentation using 2D images, the process for the construction of aligned RGB and depth pixels in the region of interest (ROI)

The process of calculating 3D coordinates uses 2D pixel locations *(u, v)* in an image, as well as a depth (d), into a 3D space, called a perspective transformation. The distance of the point *(u, v, d)* is defined as, where u and v are the locations in the image geometry, and d is the depth. We can reconstruct x and y coordinates using Eq. (1) for perspective transformation of the given depth [30]. Here, f and f are the focal lengths of the X and Y axis, which are computed from the intrinsic matrix of camera calibration [30].

$$x = \frac{uz}{fx}, \quad y = \frac{vz}{fy}, \quad z = \frac{d}{\sqrt{1 + \left(\frac{u}{fx}\right)^2 + \left(\frac{v}{fy}\right)^2}} \tag{1}$$

The depth and color key points are achieved by segmenting and aligning pixels from RGB-D data and projecting them into 3D space. The key steps performed in this research are explained below in the form of simple steps.

The RGB-D data should be aligned to produce 3D pixels exactly as they are in the experiment. The algorithm 1 explains the process of segmentation on RGB-D data. The same process was applied to both RGB and Depth images, but with better results. The RGB-D camera sensors usually include some extra pixels due to environmental noise or light factors that are not meant to be included in 3D space. To tackle this problem, a threshold was introduced with the mean distance of depth pixel values.

---

**Algorithm 1:** 3D instance segmentation from indoor environment using RGB-D data

**Input:** {RGB-D aligned image. RGB Color Image i.e., $S_R$, Depth Image i.e., $S_D$}
**Output:** {3D instance segmentation. Segmented 3D Objects, Segmented 3D Background}
1. **Initialize** $S_M$: = Mask R-CNN + PointRend ($S_R$) i.e., 2D Segmented image(s)
2. **Set** Threshold: = moderate i.e., to avoid unecessory pixels
3. **While** $S_i <= S_M$ **do**
    4. MaskedObjectArea: = FindContour($S_i$) i.e., identify objects area
    5. **if** Segment: = Objects **then**

(Continued)

---

**Algorithm 1:** Continued

---

      6. DepthObjectArea: = SD–Threshold (Masked Object Area)
      7. ColoredObjectArea: = SR–MaskedObjectArea
      8. PCD: = CreatePointCloud (
                               Colored Object Area, Depth Object Area, Camera Intrinsic)
      9. Save Segmented 3D Objects as PCD
   10. **else if** Segment: = Background **then**
      11. DepthObjectArea: = Threshold(MaskedObjectArea)-$S_D$
      12. ColoredObjectArea: = MaskedObjectArea-$S_R$
      13. PCD: = CreatePointCloud(
                               ColoredObjectArea, DepthObjectArea, CameraIntrinsic)
      14. Save Segmented 3D Background as Point Cloud
    15. **end if**
**end while**

---

## 4  Experimental Results

In a controlled setup, the Intel RealSense D-415 sensor is used for capturing RGB-D images, with the sensor positioned 3–4 meters away from objects in the indoor environment. The Intel RealSense D415 uses stereo scans with a pair of depth sensors, RGB sensors, and infrared radiation (IR) projection systems for depth estimation. The experiment revolves around the combination of a few technologies used to segment 3D objects from a 3D point cloud scene for an indoor environment. In order to project 3D points calculated after RGB-D segmentation, the Open3D was utilized. Open3D is an open-source library that supports 3D data processing and applications. It is used to build and compile from the source on various platforms with minimal effort.

In order to achieve the detection and masked objects, the Mask R-CNN deep learning model was used with the common objects in context (COCO) dataset. Microsoft released the MS COCO dataset, which is large-scale object detection, segmentation, and captioning dataset. The COCO dataset is a large-scale dataset that provides dense pixel-wise contextual annotations to images of high complexity. COCO is a tool for studying thing-thing interactions, and it includes images of complicated scenes with a lot of small items that are labeled with very detailed outlines. To address the annotations and high level of correctness, the COCO dataset contains 164 K images of 91 classes, divided into four subgroups: training (118 K), validation (5 K), test (20 K), and test-challenge (20 K) (20 K) [31].

The contribution and the findings of this research are as follows in the field of 3D instance segmentation. In Fig. 6, some findings from real-world indoor scenes without recognized objects (as a background) and separated recognized objects are projected in 3D space. The first two rows are projecting RGB-D data; the third row is projecting the first phase of the research as masked images after recognizing the targeted objects. In Fig. 6, fourth row projected as point-cloud in 3D space, generated from RGB-D data. The fifth and sixth rows represent the second phase of the research after segmenting RGB-D data into separate 3D point cloud objects and their corresponding backgrounds. Finally, all 3D matching key points are projected into 3D space with the support of the Open3D library. Finally, the process of 3d instance segmentation can be seen in Fig. 6 from top to bottom in individual column handling single and multi-objects in indoor environment.

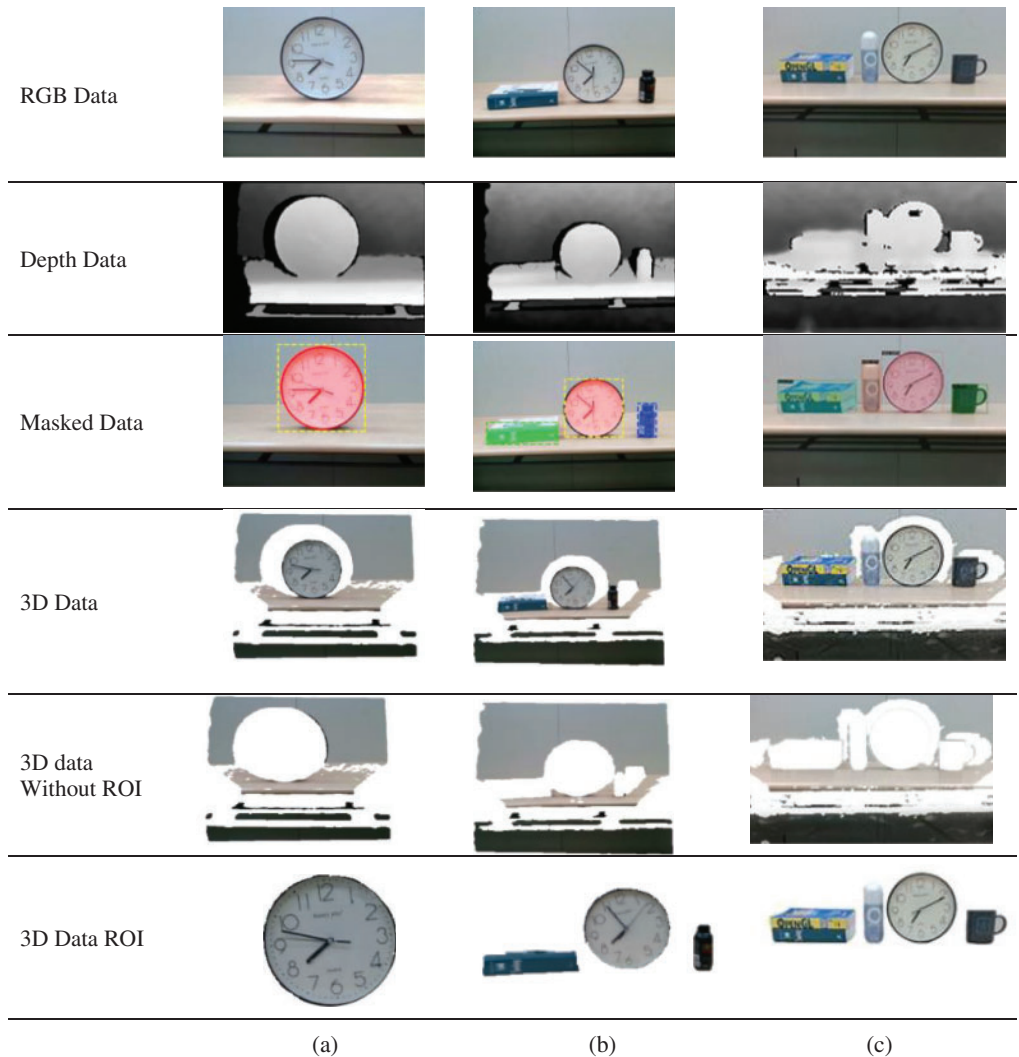

**Figure 6:** The aim of this research is to classify and segment objects from their context in 3D space using RGB-D segmentation. The distribution of columns is defined as an RGB image (First Row). The corresponding aligned depth image (Second Row). Masked area of an object using Mask-RCNN (Third Row). 3D view generated with RGB-D images (Third Row). 3D view of indoor environment without the objects (Fourth Row). The last row shows 3D view of Multi-objects without indoor (background) environment, following the process from top to bottom

The proposed method was evaluated on a PC with an NVIDIA GeForce 1080Ti GPU and an Intel® Core i7 CPU. All the images were captured with the Intel RealSense D415 camera. The RGB-D data from the first row is captured at two matters of distance from the camera; the second row is three, and the third row is captured four meters away from the camera. The RGB-D data is used to segment objects that are placed at different viewpoints.

In this research, the aim was to segment 3D instances of targeted objects and to separate them from their background within a given indoor environment using RGB-D data. To fully recognize and separate a 3D instance, one would require both a semantic label and an instance label. We adopted a simple approach to segment and recognize instances in RGB images first and then generate and separate 3D instances from their background according to 2D instance segmentation. We benefit from the real distances in 3D scenes, where the sizes and distances between objects are vital to ultimate instance segmentation. Our approach consistently outperforms 3D object recognition and instance segmentation with background separation across all three difficulty levels. Difficulty levels are considered in the form of single and multi-objects in an indoor environment. The result is adequate and clean, which is suitable for simulations and real-time applications.

The goal of detecting and segmenting simple objects in Fig. 6 was to study whether the proposed approach can recognize, segment, and separate various objects in 3D indoor environments, although their shapes are rather similar. The image (c) in Fig. 6 illustrates how the objects are segmented into a 2D RGB image, where different colors represent separate target instances. The method depicted in Fig. 3 Select each masked instance from the scene so that it can be processed later. Each distinct identified object may be segmented individually by applying a technique explained in algorithm Section 3.3 by considering individual masked segmented images. Finally, we can visualize the 3D environment in Fig. 6, and 3D instances are effectively extracted separately from the 3D point cloud. In Fig. 6 we visualize representative outputs of our proposed method. We see that for simple cases of non-occluded objects at a reasonable distance (so we get a sufficient number of points), our approach outputs remarkably accurately in 3D instance segmentation and separate their background. For 3D instance segmentation with different or nearby or occluded objects, the result of mixed segmentation appears to be very challenging in some cases. Because our method is based on the Mask R-CNN and PointRend 2D instance segmentation models, we could potentially mitigate this problem if we propose a solution in future.

The quantitative assessment of real-world scenes is challenging since no ground-truth model is suitable for this specific scenario. A comprehensive comparison of all segmented objects in each scene reconstituted 3D instances has been collected in Fig. 6. The exactness of our proposed method is demonstrated by comparing the average width measurements of segmented 3D objects to real-world original measurements in Tab. 2.

**Table 2:** A controlled assessment in millimeters of 3D instance and an object from the real world is provided. From left to right in Fig. 6 Column (c)

| Objects | Measurements | 3D Measurements | Error |
|---|---|---|---|
| Book (width) | 203 mm | 200 m | 3 mm |
| Bottle (height) | 126 mm | 124 mm | 2 mm |
| Clock (diameter) | 228 mm | 227 mm | 1 mm |
| Cup (width) | 76 mm | 74 mm | 2 mm |

The calculated difference clearly explains the accuracy of our proposed method between segmented 3D objects and real-world objects in the last column of Tab. 2. The calculated difference is nominal against every object, which can be neglected by different applications used for various systems, such as intelligent and robot systems.

Currently, five objects are used to segment for experiment in one frame of RGB-D data. Width and height are calculated between different objects for observing the calculated difference between real world and 3D segmented objects. Due to nominal measurement difference in real world and segmented objects the 3D instance segmentation presented in this paper for RGB-D data that can be deployed on a variety of systems with minimal effort and resources.

In Fig. 6, we demonstrate the outcomes of our suggested 3D instance segmentation method employing RGB-D data, as well as the margin error in Tab. 2. The method we present successfully distinguishes between many objects and their surroundings. To the best of our knowledge, there is currently no way to separate objects from their backgrounds using segmentation. Most of the proposed approaches can segment items in a scene, but none of them can separate objects from their backgrounds. Many investigations employ a specific dataset for 3D objects without considering their context or background; however, our study focuses on segmenting objects and their backgrounds as two distinct outcomes. As robotics applications require information about their surrounds and context to the items, such an output is useful for intelligent robot system applications. As demonstrated in Tab. 2, the error margin is quite small in our scenario, (1–3 millimeters) which is nominal for the movement or grasping of robot systems. In addition, as a follow-up to our current research, we want to conduct some experiments and comparisons with some existing methods after making structural adjustments to our method to match our output with that of existing methods. Another future project is to segment under some obstructed or congested conditions with a 360-degree view of the surroundings. As a result, segmented object measurements are critical for intelligent and robot system applications to overcome margins of error in real-world settings, as demonstrated in this study.

## 5 Conclusion

In this paper, a novel approach to 3D instance segmentation and detachment of multi-objects and their context from the indoor environment has been presented, that has not been explored before. The approach uses RGB-D data from the Intel RealSense sensor and recognizes targeted objects in RGB data by utilizing the PointRend with Mask R-CNN. An effective mechanism has been developed and evaluated for 3D instance segmentation and detaching the 3D instances from their background. The method of segmenting 3D instances is achieved by identifying masked pixels in RGB and depth data. Furthermore, a comprehensive comparison of the proposed 3D instance segmentation approach for the indoor environment was collected and accessed. The outcome of this research focused on 3D instance segmentation for single as well as multi-objects from an indoor environment. The same approach can be applied to multiple overlapping RGB-D data frames by utilizing the registration approaches.

Research has resulted in the effectiveness of segmentation and separation of objects in an indoor environment for robot vision. Different applications may use separated objects and their backgrounds, i.e., intelligent and robot systems. Since robots use precise xyz coordinates to grasp objects, our proposed approach has a small discrepancy between actual measurements and 3D object measurements that robot systems may neglect.

Although the research output is adequate, certain concerns in the proposed approach still have the potential for improvements. For instance, in the identification of occluded objects and their 3D instance segmentation, noise may occur due to different camera sensors and environmental noise in the 3D view of scenes. Furthermore, the research intends to bring this work forward to the next stage involving the registration 360° degree process in multiple distinct overlapping scenes. After obtaining a registered view of several overlapping RGB-D data frames, 3D instances can be segmented and separated from their background in 360°.

**Funding Statement:** This research was supported by the BB21 plus funded by Busan Metropolitan City and Busan Institute for Talent & Lifelong Education (BIT) and a grant from Tongmyong University Innovated University Research Park (I-URP) funded by Busan Metropolitan City, Republic of Korea.

**Conflicts of Interest:** The authors declare no conflict of interest.